\documentclass[runningheads]{llncs}
\usepackage{booktabs, multirow}
\usepackage{graphicx}
\usepackage{amsfonts, amsmath}
\usepackage{floatflt}
\usepackage{caption}
\usepackage{subcaption}
\newcommand{\half}{\frac{1}{2}}
\newcommand{\fhat}{\hat{f}}

\begin{document}
\title{Multi-Agent Learning of Numerical Methods for Hyperbolic PDEs with Factored Dec-MDP\thanks{This work was supported in part by the Defense Advanced Research Projects Agency (DARPA) under Agreement No. HR00112090063.
}}
%
%
\author{Yiwei Fu\inst{1} \and 
Dheeraj S.K. Kapilavai\inst{1} 
\and Elliot Way\inst{2}
}
\authorrunning{Y. Fu et al.}
\titlerunning{Multi-Agent Learning of Numerical Methods with Factored Dec-MDP}
%
\institute{
{GE Research, Niskayuna NY 12309, USA} \email{\{yiwei.fu, kapilava\}@ge.com}
\and {Binghamton University, Binghamton, NY 13902, USA \email{ellioteway@gmail.com}}
}

\maketitle              
\begin{abstract}
Factored decentralized Markov decision process (Dec-MDP) is a framework for modeling sequential decision making problems in multi-agent systems.
In this paper, we formalize the learning of numerical methods for hyperbolic partial differential equations (PDEs), specifically the Weighted Essentially Non-Oscillatory (WENO) scheme, as a factored Dec-MDP problem.
We show that different reward formulations lead to either reinforcement learning (RL) or behavior cloning, and a homogeneous policy could be learned for all agents under the RL formulation with a policy gradient algorithm.
Because the trained agents only act on their local observations, the multi-agent system can be used as a general numerical method for hyperbolic PDEs and generalize to different spatial discretizations, episode lengths, dimensions, and even equation types.

\keywords{Decentralized Markov Decision Process \and Multi-Agent System \and Numerical Method \and Partial Differential Equation}
\end{abstract}
\section{Introduction}

Numerical methods such as the finite difference method (FDM), finite element method (FEM), and finite-volume method (FVM) were designed on grids of data points to approximate the solutions of partial differential equations (PDEs) numerically. 
Specifically, hyperbolic PDEs are often used in real-world applications, especially in the field of fluid dynamics for modeling the motion of viscous fluids with high Reynolds numbers, multiphase flows, water waves, etc. 
Often, simulations of these physical processes exploit the numerical fluxes of some conserved quantities from finite differences between two elements, and time integration is used to determine the next state.
This kind of incremental computation in time is similar to a Markov Decision Process (MDP), where the next state is determined only by the current state, and actions provided by the numerical method.
Since numerical computations happen at all grid points simultaneously, this can be viewed as a multi-agent system where each agent only has partial knowledge of the state, modeled by a decentralized MDP (Dec-MDP)~\cite{beynier2013dec}.

The reason that a Dec-MDP is the proper framework for modeling numerical methods can be justified from two perspectives.
On one hand, if we let a single agent act on the entire physical space, which often has at least hundreds of discretized locations and requires multiple actions at each location, the number of action dimensions would be huge, with each one being continuous.
Such large continuous action space quickly becomes impossible to deal with.
Even with the most recent advances in deep Reinforcement Learning (RL), DQN~\cite{mnih2013playing} and all of its follow-ups are only able to handle large observation spaces, while the action spaces remain small discrete ones.
Many Deep RL applications in robotics and control~\cite{lillicrap2015continuous} do handle continuous action spaces, but they usually have less than ten dimensions, because there are simply not that many degrees of freedom in one robot.
Besides, learning a numerical method with a single agent is not generalizable: any changes in spatial discretization require retraining.

On the other hand, existing numerical schemes cannot be treated as multiple independent agents acting alone at their respective locations.
PDEs that describe some underlying processes must have physics that propagates from one location to the other, so if the action at one location changes, the state at another location would be different even if the action at that location stays the same.
This causes the non-stationary issue in a multi-agent environment~\cite{zhang2021multi}.
Therefore, to properly model numerical methods, a Dec-MDP framework has to be used to account for the effects of other agents.

Specifically in this paper, we focus on learning Weighted Essentially Non-Oscillatory (WENO)~\cite{Liu1994WeightedEN,shu1998essentially}, a state-of-the-art numerical scheme with a uniform high order of accuracy in flux reconstruction.
The main idea of WENO is to form a weighted combination of several local reconstructions based on different stencils and use it as the final WENO reconstruction.
The combination coefficients (weights) depend on the linear weights and smooth indicators~\cite{zhang2016eno}.
Later we will highlight the insight that these weights can be viewed as actions of agents. 

In the field of multi-agent systems, it is well known that optimally solving a Decentralized Partially Observable Markov Decision Process (Dec-POMDP) is NEXP-complete~\cite{bernstein2002complexity}.
Although Dec-POMDP is a more generalized formulation of Dec-MDP, with the latter being the jointly fully observable version of the former, Dec-MDP is still NEXP-complete. 
In our previous work~\cite{way2022backpropagation}, a policy gradient algorithm called Backpropagation Through Time and Space (BPTTS) is proposed to solve the multi-agent reinforcement learning problem for numerical methods with recurrent neural networks (RNNs). 
Here we use BPTTS to solve a factored Dec-MDP problem, and show that automatic differentiation is possible for the RL formulation to learn a generalizable numerical scheme.

In this paper, we introduce the fundamentals of both Dec-MDPs and numerical methods.
Then, the key insight that the learning of numerical methods can be modeled as a factored Dec-MDP is drawn.
We analyzed different reward formulations which lead to either RL or behavior cloning.
A policy gradient algorithm, BPTTS~\cite{way2022backpropagation}, is used to solve the homogeneous multi-agent RL problem. 
We show that the learned policy can generate numerical solutions for hyperbolic PDEs comparable to WENO methods, and can generalize to different grid discretizations, initial conditions, dimensions, and equations.
This practical application of using a factored Dec-MDP framework to solve hyperbolic PDEs is novel, and could potentially have real impacts on both industry and academia.

\section{Related Work}~\label{sec:related_work}

In the field of numerical analysis, there are two distinct classes of methods: mesh-based methods and meshfree methods. 
The comparison of these two is beyond the scope of this paper, but usually they are incompatible, thus so are the machine learning approaches that try to learn them.
For example, the WENO method~\cite{shu1998essentially} used in this paper is a mesh-based method; while on the other hand, the popular Physics Informed Neural Network (PINN)~\cite{raissi2019physics} and their descendants are meshfree: they directly use neural networks to approximate the solutions, gradients, or operators of PDEs.
Since both the spatial and temporal coordinates are explicit in PINN inputs, modern deep learning architectures which handle spatial or temporal information implicitly, like Convolutional Neural Networks (CNNs), Recurrent Neural Networks (RNNs), etc., are not directly applicable.
As a result, PINNs often resort to fully-connected neural networks.

Among numerous papers at the intersection of machine learning and mesh-based numerical methods, the setup in Wang et al.~\cite{wang2019learning} resembles that of this paper most closely.
Although they are also trying to learn a high-order numerical scheme for hyperbolic PDEs, they treat this inherently multi-agent problem as multiple independent single-agents.
This will lead to a sub-optimal solution at best because for each agent the environment becomes non-stationary: the actions taken by one agent will affect the rewards of other agents and the state evolution at other locations.
This invalidates the stationary Markovian assumption 
that the individual reward and current state should depend only on the previous state and actions taken.
There exist some other papers that claim to have used multi-agent RL to discover closure models in simulations of turbulent flows~\cite{novati2021automating,bae2022scientific}.
However, they are not formulated under a truly decentralized framework because of the use of replay buffers.
When sampling different state-action pairs inside a replay buffer, the temporal correlations are broken, therefore the effects of other agents' actions over time on one agent's current state are lost, again leading to a non-stationary environment.
To sum up, solutions to these multi-agent systems must be consistent both spatially and temporally.

In the field of multi-agent systems, there have been some previous papers on factored Dec-POMDPs, which is close to the formulation in this paper. 
Oliehoek et al.~\cite{oliehoek2008exploiting} exploit the locality of interactions between agents in a factored Dec-POMDP and formulated decomposable value functions. 
This leads to a single framework based on collaborative graphical Bayesian games and is solved by heuristic policy search. 
However, this approach is only tested on a simple factored Firefighting problem (FFP) with three agents.
Later on, this was extended to deal with more agents~\cite{oliehoek2013approximate} by using a factored forward-sweep policy computation that tackles the stages of the problem one by one, but the action space for each agent remains the same (two discrete choices). 
Pajarinen et al.~\cite{pajarinen2011efficient} propose an expectation-maximization based optimization for factored infinite-horizon Dec-POMDP and kept the complexity tractable by factored approximations. 
They apply the algorithm to the same FFP with a maximum of $10$ agents. 
Messias et al.~\cite{messias2011efficient} convert a factored Dec-POMDP to a centralized Multi-agent POMDP by allowing inter-agent communications.
Amato et al.~\cite{amato2014planning} model macro-actions as options in a factored Dec-POMDP model to model systems where coordination decisions only occur at the level of deciding which macro-actions to execute.
They have demonstrated that near-optimal solutions can be generated for longer horizons and larger state spaces than previous Dec-POMDP methods.

\section{Background}\label{sec:background}
In this section, we introduce the background of both factored Dec-MDP for modeling multi-agent systems, and weighted essentially nonoscillatory (WENO) scheme, a high-order accurate numerical scheme for hyperbolic PDEs.

\subsection{Factored Dec-MDP}\label{sec:background:dec-mdp}
Before defining the multi-agent framework used in this paper, factored Dec-MDP, it is helpful to first introduce Dec-POMDP and Dec-MDP, because factored Dec-MDP is a special case of Dec-MDP, who is a special case of Dec-POMDP~\cite{beynier2013dec}.

\begin{definition}~\label{defn:dec_pomdp}
\textbf{Dec-POMDP:} A Dec-POMDP for $n$ agents is defined as a tuple $\langle\mathcal{S}, \mathcal{A}, \mathcal{P}, \Omega, \mathcal{O}, \mathcal{R}\rangle$ where:
\begin{itemize}
	\item $\mathcal{S}$ is a finite set of system states;
	\item $\mathcal{A}=\left\langle\mathcal{A}_{1}, \ldots, \mathcal{A}_{n}\right\rangle$ is a set of joint actions; $\mathcal{A}_{i}$ is the set of actions $a_{i}$ that can be executed by agent $\mathcal{A}g_{i}$;
	\item $\mathcal{P}=\mathcal{S} \times \mathcal{A} \times \mathcal{S} \rightarrow[0,1]$ is a transition function; $\mathcal{P}\left(s, a, s^{\prime}\right)$ is the probability of the outcome state $s^{\prime}$ when the agents execute the joint action $a$ from $s$;
	\item $\Omega=\Omega_{1} \times \Omega_{2} \times \cdots \times \Omega_{n}$ is a finite set of observations, where $\Omega_{i}$ is $\mathcal{A}g_{i}$'s set of observations;
	\item $\mathcal{O}=\mathcal{S} \times \mathcal{A} \times \mathcal{S} \times \Omega \rightarrow[0,1]$ is the observation function; $\mathcal{O}\left(s, a, s^{\prime}, o=\right.$ $\left.\left\langle o_{1}, \ldots, o_{n}\right\rangle\right)$ is the probability that each agent $\mathcal{A} g_{i}$ observes $o_{i}$ when they execute the joint action $a$ from state $s$ and the system moves to state $s^{\prime}$
	\item $\mathcal{R}$ is a reward function; $\mathcal{R}\left(s, a, s^{\prime}\right)$ is the reward the system obtains when the agents execute joint action a from state $s$ and the system moves to state $s^{\prime}$.
\end{itemize}
\end{definition}


\begin{definition}~\label{defn:dec_mdp}
	\textbf{Dec-MDP:} A Dec-MDP is a special case of Dec-POMDP where the system state is jointly observable, i.e.:
	\begin{itemize}
		\item If $\mathcal{O}(s,a,s^{\prime},o=\langle o_1,...,o_n \rangle)>0$, then $Pr(s^{\prime}|\langle o_1,...,o_n \rangle)=1$.
	\end{itemize}
\end{definition}
Note that joint observability does not entail local observability. For each individual agent, the full system state is still partially observable.

\begin{definition}~\label{defn:factored_dec_mdp}
	\textbf{Factored Dec-MDP:} A factored Dec-MDP is a Dec-MDP where the state of the system $\mathcal{S}=\mathcal{X}_1 \times \mathcal{X}_2 \times \cdots \times \mathcal{X}_{|\mathcal{X}|}$ has $|\mathcal{X}|$ components and is spanned by $\mathcal{X}=\{\mathcal{X}_1, \mathcal{X}_2, \cdots, \mathcal{X}_{|\mathcal{X}|}\}$. 
\end{definition}
Like factored Dec-POMDP in~\cite{oliehoek2008exploiting}, the reward function of factored Dec-MDP can often be compactly represented by exploiting \textit{additive separability}, meaning that the total reward can be decomposed into the sum of local reward functions $R=R^1+\cdots+R^p$. The local reward functions are often defined over a smaller number of state and action variables and the scope of them is smaller.

\subsection{Hyperbolic PDEs and WENO Scheme}\label{sec:background:weno}
Conservation laws in many branches of classical physics, such as fluid dynamics and electrodynamics, are often described by hyperbolic PDEs.
The finite-difference grid is often used when dealing with such equations.
The goal is to evolve a vector of conserved quantities $u \in \mathbb{R}^d$ on a uniform $N$-point discretization $D_N=\{x_1, x_2,\ldots,x_j,\ldots, x_N\}$ for hyperbolic PDE of the form: 
\begin{equation}~\label{eqn:pde}
\frac{\partial u}{\partial t} + \frac{\partial}{\partial x}f(u) = 0
\end{equation}
where $f(u)$ are fluxes of each quantity in $u$ that are exchanged at the interface $x_{j\pm\half}$ of each cell, $I_j= [x_{j-\half},x_{j+\half}]$. 
The initial conditions, $u(x_j,0)$ at all locations at the beginning of the simulation, along with boundary conditions, $u(x_1, t)$ and $u(x_N, t)$ at all times, are required to specify the PDE. 
Then, the method of lines~\cite{schiesser2012numerical} is used to convert the PDE to an ordinary differential equation (ODE), where the spatial derivative is approximated by the finite differences:
\begin{equation}
\label{eqn:fd}
\frac{d u_j(t)}{d t} = - \frac{1}{\Delta x}(\fhat_{j+\half} - \fhat_{j-\half})
\end{equation}
with $u_j$ being the approximation to the point value $u(x_j,t)$ and $\fhat_{j\pm\half}$ being the numerical fluxes computed at the interfaces using cell values. 
A high-order finite difference scheme boils down to
\begin{itemize}
    \item using a high-order time integration scheme (e.g., a high-order Runge-Kutta method);
    \item using the finite difference in Equation~(\ref{eqn:fd}) to approximate the derivative of the flux to a high order.
\end{itemize}
WENO scheme~\cite{shu1998essentially} is one high-order scheme. 
A WENO scheme of order $r$ can achieve a $2r-1$ order spatially accurate construction of $\fhat_{j+\half}$ by using the stencil $S_j$ with $2r-1$ points around $u(x_j,t)$. 
Specifically, it computes $\fhat_{j+\half}$ as a convex combination of polynomials defined on $r$ small stencils inside $S_j$:
\begin{equation}
\fhat_{j+\half} =\sum_{k=0}^{r-1} \omega_k \fhat_{k,j+\half} \quad \text{with} \quad \sum_{k=0}^{r-1} \omega_k = 1
\label{eqn:weno_weights}
\end{equation}
where $\fhat_{k,j+\half}$ is the polynomial reconstruction of the $k$-th small stencil.
For example, for a $2$-nd order WENO scheme at location $x_j$, the stencil is:
\begin{equation}
S_j = \{x_{j-1}, x_{j}, x_{j+1}\} \\
\label{eqn:stencils}
\end{equation}
and the $2$ small stencils are $\{x_{j-1}, x_{j}\}$, $\{x_{j}, x_{j+1}\}$ respectively.


To choose these convex weights $\omega_k$ and ensure they sum up to $1$, WENO scheme computes the following:
\begin{equation}
\omega_{k}=\frac{\alpha_{k}}{\sum_{m=1}^{r} \alpha_{m}} \quad \text{with} \quad \alpha_{k}=\frac{d_{k}}{\left(\varepsilon+\beta_{k}\right)^{2}}
\label{eqn:weno_scheme}
\end{equation}
where $d_k$ is a pre-computed optimal coefficient determined by a smoothness indicator $\beta_k$, and $\epsilon$ is a small positive number to avoid the denominator becoming zero.
In smooth regions, these weights are designed to produce higher-order approximations.
At discontinuities, WENO can select the single best small stencil to avoid discontinuities as far as possible by setting the one $\omega_k$ corresponding to that small stencil to be $1$, and all other weights to be $0$.
Since the core of WENO scheme is actually an approximation procedure not directly related to hyperbolic PDEs or finite difference methods, it can also be generalized and applied to other types of schemes (finite volume, compact schemes, residual distribution schemes, limiters for the discontinuous Galerkin schemes, etc.), or to different fields~\cite{shu2009high}.

\section{Problem Formulation and Analysis}~\label{sec:formulation}
With the background information provided in Section~\ref{sec:background}, here we draw the connection between numerical methods and factored Dec-MDP, and provide some analysis on the reward formulations.

\subsection{Numerical Methods as Multi-Agent Systems}~\label{sec:formulation:numerical}
\begin{figure}[h]
  \centering
  \includegraphics[width=0.8\textwidth]{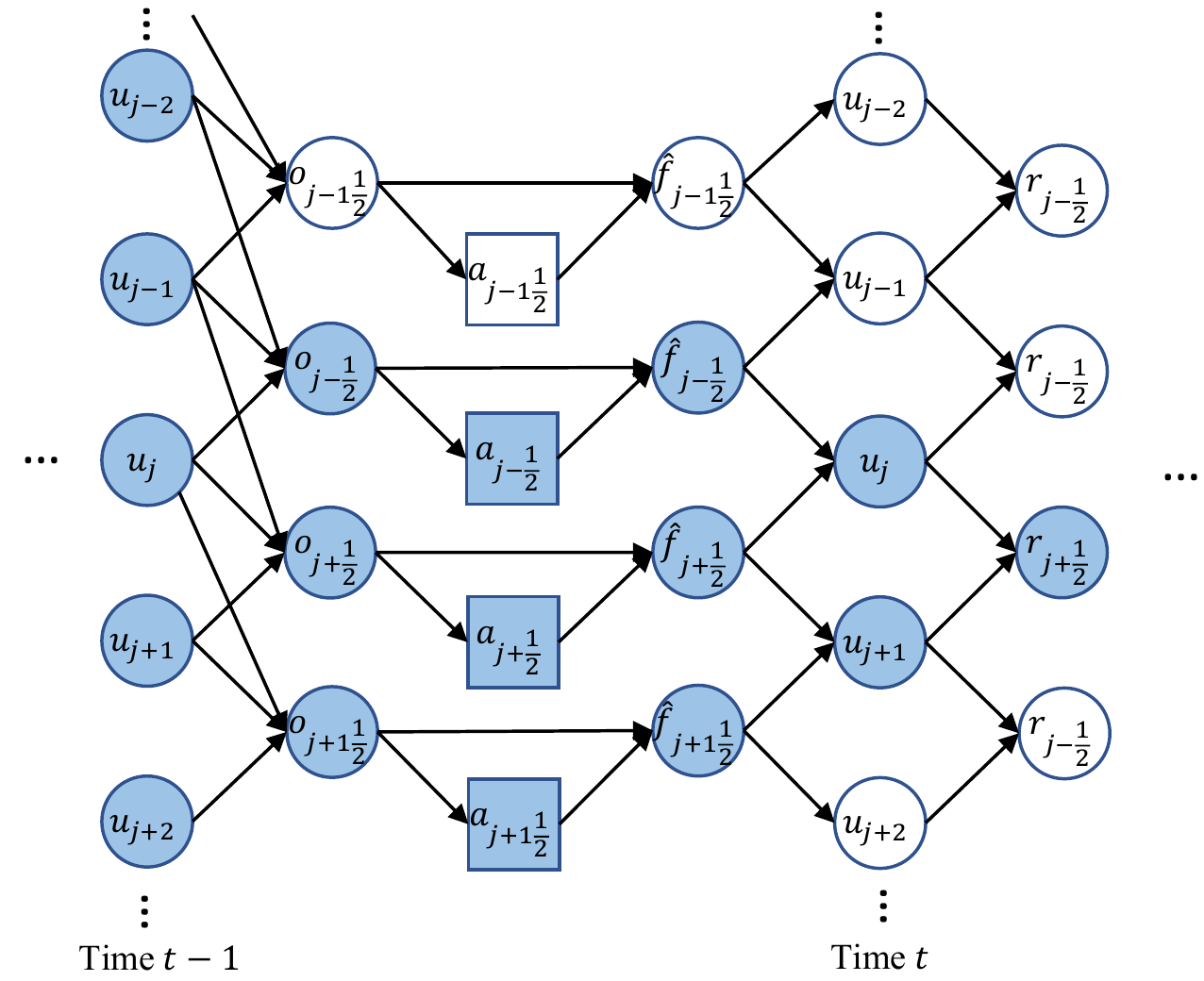}
  \caption{The dynamics of the numerical scheme over two steps. The scope of $r_{j+\frac{1}{2}}$, illustrated by shading, increases when going back in time.}
  \label{fig:scope}
\end{figure}
Because of the difficulty of translating a PDE into a computable model, the design of numerical schemes often requires substantial efforts by domain experts.
If the discovery of numerical schemes can be automated, 
it could potentially have huge impacts on many applications. 
We make the key connection here that WENO scheme works exactly like a collaborative multi-agent system.

To formulate the learning of numerical methods for hyperbolic PDEs as a factored Dec-MDP problem, we can start with the discretization of the physical space in Equation~(\ref{eqn:pde}), which is precisely factoring the entire system state  $\mathcal{S} \equiv u$ into $n$ components $u_1 \times u_2 \cdots \times u_n$ as in Definition~\ref{defn:factored_dec_mdp}. 
From Equation~(\ref{eqn:weno_weights}) to (\ref{eqn:weno_scheme}), we can see that for WENO scheme at a certain location, it is essentially doing a local observation on the stencil and then coming up with the weights to recompute the fluxes.
This process is exactly an agent taking an action $a$ after observing a state, with the action being the weights $\omega_k$ in Equation~(\ref{eqn:weno_weights}). 
Then, the PDE (or environment) uses Equation~(\ref{eqn:fd}) to integrate the system to the next state, which defines a deterministic transition function. 
This system state is Markovian and the next set of agents' actions only depends on the next state, because WENO scheme does not keep track of the history. 
Furthermore, since the joint observation for all agents fully covers the entire physical space, the system state is jointly observable, but not locally fully observable. 
Therefore, the learning of WENO schemes can be properly modeled as a factored Dec-MDP problem.

In order to learn WENO scheme, the immediate reward (the details of which will be discussed in Section~\ref{sec:formulation:reward}) is defined at interface $j+\frac{1}{2}$ as the average of the error in its two adjacent cells. 
An error of $0$ means that our agents have learned actions similar to WENO actions so that the PDE system state evolves to the same one. 
Following the definition of scope in~\cite{oliehoek2008exploiting}, we can draw the dynamics of the numerical scheme in Figure~\ref{fig:scope} and observe that the scope increases when going back in time.
This should make sense from both the multi-agent system perspective and the physics perspective: the action of an agent at a certain time and space would affect the observations of other agents at a later time, or the physics would propagate from one location to another.
This is precisely the reason that a multi-agent system should not be modeled as multiple independent single agents as in~\cite{wang2019learning}.
It should be noted that as with our previous work~\cite{way2022backpropagation}, Lax-Friedrichs flux splitting~\cite{lax1954weak} was used to ensure numerical stability and avoid entropy-violating solutions, which makes the flux reconstruction slightly different by splitting them into plus and minus terms, but the main idea still holds.

\subsection{Analysis of Different Reward Formulations}~\label{sec:formulation:reward}
Among components of the factored Dec-MDP, the transition function and observation function are already determined by physics, so the only moving piece is the reward function.
We formalize the reward at interface $j+\frac{1}{2}$ as follows:
\begin{equation}~\label{eqn:reward}
    r^t_{j+\frac{1}{2}}=-\frac{|u^t_j - ref^t_j| + |u^t_{j+1} - ref^t_{j+1}|}{2}
\end{equation}
where $u^t_j$ is factored state $u_j$ at time $t$, and $ref^t_j$ is a reference state at the same location $j$ and time $t$ computed by some existing numerical methods like WENO. 
Maximizing this reward will lead to the agents trying to evolve the state to be as close to the reference state as possible.
Because the system immediate reward at time $t$ is the summation of $r^t_{j+\frac{1}{2}}$ in Equation~(\ref{eqn:reward}) at all interfaces, $r^t=\sum_j r^t_{j+\frac{1}{2}}=-\sum_j |u^t_j - ref^t_j|$ (with a small caveat that the boundary conditions needed to be computed according to the physics), we can further analyze the reward function with integration from Equation~(\ref{eqn:fd}) and (\ref{eqn:weno_weights}) as follows:
\begin{equation}~\label{eqn:analyze_reward}
\begin{aligned}
    r_t &= -\sum_j |u^t_j - ref^t_j| = -\sum_j |u_j^{t-1} - \frac{\Delta t}{\Delta x}(\fhat_{j+\half} - \fhat_{j-\half}) - ref^t_j| \\
    &= -\sum_j | u_j^{t-1} - \frac{\Delta t}{\Delta x}\sum_{k=0}^{r-1} (a_{k,j+\half} \fhat_{k,j+\half} - a_{k,j-\half} \fhat_{k,j-\half}) - ref^t_j|
\end{aligned}
\end{equation}

There are different choices of reference states, and different reward functions can be combined.
However, reward engineering is beyond the scope of this paper, here we analyze $3$ simple reward formulations for the PDE environment:
\begin{enumerate}
    \item \textit{RL-WENO} (Markovian WENO agent-based simulation rewards): the reference state is calculated from $u^{t-1}$ with the standard WENO scheme following the computational graph in Figure~\ref{fig:scope}. This reward is Markovian because the actions of the standard WENO actions depend only on the most recent state. This reward leads to a \textit{reinforcement learning} problem. Equation~(\ref{eqn:analyze_reward}) becomes $r_t = -\frac{\Delta t}{\Delta x}\sum_j \sum_{k=0}^{r-1}|(a_{k,j+\half} - \omega_{k,j+\half}) \fhat_{k,j+\half} - (a_{k,j-\half}-\omega_{k,j-\half}) \fhat_{k,j-\half})|$, with $\omega_{k}$ being the actions taken by the WENO scheme at $u_k^{t-1}$.
    \item \textit{BC-WENO} (Non-Markovian fixed WENO solution rewards): the reference state is pre-calculated by running a WENO scheme from an initial condition all the way till the end. This reward is non-Markovian because the reference state is fixed and does not depend on the current system state. This fixed expert trajectory leads to \textit{behavior cloning} and is a \textit{supervised learning} problem. Equation~(\ref{eqn:analyze_reward}) becomes $r_t = -\frac{\Delta t}{\Delta x}\sum_j \sum_{k=0}^{r-1}|a_{k,j+\half} \fhat_{k,j+\half} - C_{k,j+\half}^t - a_{k,j-\half} \fhat_{k,j-\half} + C_{k,j-\half}^t|$, with $C_{k,j+\half}^t$ and $C_{k,j-\half}^t$ being constants predetermined by the reference state trajectory.
    \item \textit{BC-analytical} (Non-Markovian fixed analytical solution rewards): similar to \textit{BC-WENO}, but the reference trajectory is pre-calculated by using the analytical solution to the PDE. Of course, this requires the existence of such analytical solutions to begin with. This reward also leads to \textit{behavior cloning}. The reward structure is similar to \textit{BC-WENO}, but with different constants calculated by the analytical solution.
\end{enumerate}

With the reward function defined and exploiting the fact that all state transitions are differentiable in this PDE environment, we could train a policy gradient algorithm (such as BPTTS~\cite{way2022backpropagation}) for this homogeneous multi-agent system to learn numerical methods. 
The policy is parameterized by a fully connected neural network (NN) whose weights are shared by all agents.
During training, gradients flow back in time and space following the shaded routes as illustrated in Figure~\ref{fig:scope} and an RNN-like computational graph is created. 
However, this recurrent structure is not used once the agents are trained, because in a factored Dec-MDP, each agent's actions only depend on its immediate observations. 
This leads to incredible generalizability for the multi-agent system: they can be applied to different spatial and/or temporal discretizations, as shown in Section~\ref{sec:exp}.

Although the performance of agents in \textit{RL-WENO} is upper-bounded by the WENO scheme (so is \textit{BC-WENO}), i.e., the best learning results would be $a_{k,j+\half} = \omega_{k,j+\half}$ or the agents take actions exactly like WENO schemes leading to the maximum reward of 0, the RL formulation provides a straightforward framework for the agents to learn. 
This is also the reward used in our previous work~\cite{way2022backpropagation}.
\textit{BC-WENO} and \textit{BC-analytical} not only suffer from the distributional drift problem in behavior cloning~\cite{ross2013interactive}, but also have an obvious local minimum where $a_{k,j+\half} \fhat_{k,j+\half} = a_{k,j-\half} \fhat_{k,j-\half}$ since $C_{k,j+\half}^t, C_{k,j-\half}^t$ are just constants. 
This could potentially make gradient descent difficult, as shown in Section~\ref{sec:exp}.

\section{Experiment Results}~\label{sec:exp}
In this section, we introduce the Euler equations as an example of hyperbolic PDEs, and compare the training on the 3 different reward formulations proposed in Section~\ref{sec:formulation:reward}.
We then show that the trained system of \textit{RL-WENO} agents can generalize to different spatial discretizations and temporal lengths, initial conditions (ICs), equations, and even 2D Euler equations.

\subsection{Euler Equations and Training Setup}~\label{sec:exp:euler}
The Euler equations describe the conservation of mass, momentum, and energy for fluids. They are given by:
\begin{equation}~\label{eqn:euler}
	\mathcal{U}_{t}+[\mathbf{F}(\mathcal{U})]_{x}=0 \quad \text{with} \quad 
	\begin{gathered}
		\mathcal{U}=\left(\begin{array}{c}
			\rho \\
			\rho u \\
			\rho E
		\end{array}\right) \quad \mathbf{F}(\mathcal{U})=\left(\begin{array}{c}
			\rho u \\
			\rho u u+p \\
			\rho u E+u p
		\end{array}\right)
	\end{gathered}
\end{equation}
where $\rho$ is the density, $u$ is the velocity, $p$ is the pressure and $E$ is the total energy. $E$ is calculated by internal energy, $e$, and kinetic energy as $E = e + \frac{1}{2}u^2$.
The equations are closed by the addition of an equation of state, a common choice of which is the gamma-law given by $p=\rho e (\gamma-1)$ where $\gamma=1.4$.

Following the training setup in~\cite{way2022backpropagation}, we trained a system of agents to learn the order $r=2$ WENO schemes on the Sod initial condition~\cite{sod1978survey}. 
During training, the space is discretized into $N=128$ points, i.e., the factored Dec-MDP has a system state $\mathcal{S}$ of 128 factors, resulting in a state of the shape $(3,128)$ for the 3 equations. 
For agents trying to learn the $r=2$ WENO scheme, the system observation space has a shape of $(3, 129, 2, 3)$: 3 equations, 129 agents for each equation, 2 plus and minus fluxes (because of Lax-Friedrichs flux splitting mentioned in Section~\ref{sec:formulation:numerical}) and 3 points in each stencil. 
The corresponding action space shape is $(3, 129, 2, 2)$: the first three dimensions are the same as the observation space, and the last is the 2 weights (actions) on the small stencils as given in Equation~\ref{eqn:weno_weights}. 
Each component of the observation and action space is continuous.

During training, the system is evolved to 1,000 timesteps of 0.0001 seconds for a total of 0.1 seconds for each episode, and a total of 10,000 episodes. 
The policy network for each agent is a 2-layer NN with 64 neurons each and ReLU activation. 
An Adam optimizer~\cite{kingma2014adam} with a learning rate of 0.0003 is used to train the policy gradient algorithm.
The training takes about 2 days on a 2.2GHz CPU (GPU was tried but turned out to be slower because most of the computations are in the environment, not during NN training).
Once trained, the system of agents can perform inference at a similar speed as standard WENO schemes.

\subsection{Results and Discussions}~\label{sec:exp:results}
\begin{figure}[h]
  \centering
  \includegraphics[width=0.75\textwidth]{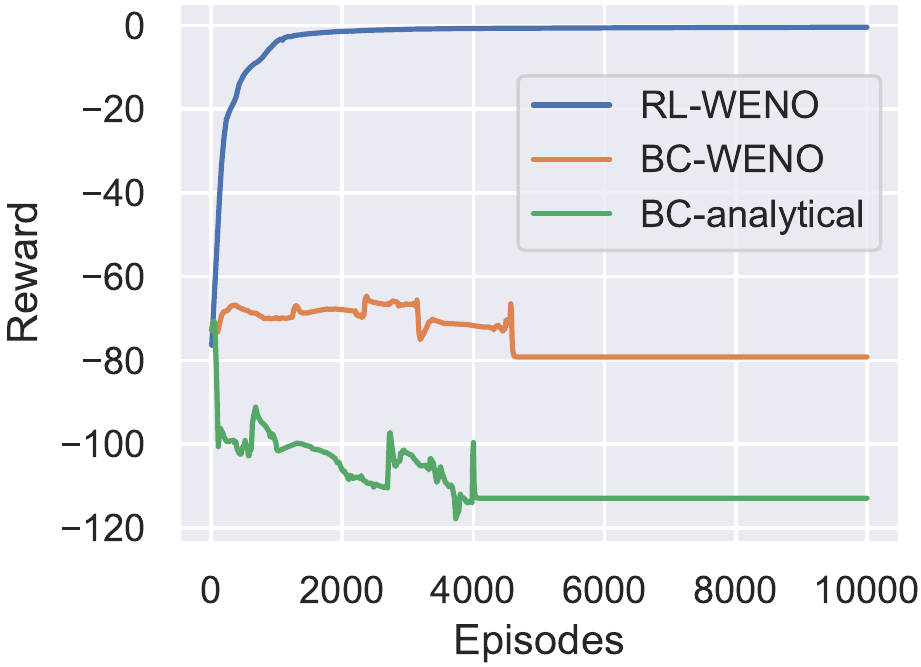}
  \caption{Reward during training for different reward formulations.}
  \label{fig:reward_comparison}
\end{figure}

The same multi-agent system is trained with the 3 reward formulations described in Section~\ref{sec:formulation:reward} and their total rewards during training are shown in Figure~\ref{fig:reward_comparison}. 
The agents are not able to learn under either \textit{BC-WENO} or \textit{BC-analytical} reward formulation (potentially trapped in a local minimum), but in contrast, \textit{RL-WENO} is successful. 
In fact, when we train agents on $r\geq3$ schemes, \textit{BC-WENO} and \textit{BC-analytical} would lead to gradient explosion immediately and the simulation has to be terminated. 
This is not to prove that they cannot be learned, but merely to show that behavior cloning can be difficult to train in this multi-agent system.
Trained \textit{RL-WENO} agents are used for the rest of this section.

We then show the generalizability of the trained \textit{RL-WENO} agents in Table~\ref{table:euler_test}. 
They were tested on different initial conditions (ICs) and evolved to longer timesteps, as detailed in~\cite{way2022backpropagation}. 
Because in this factored Dec-MDP formulation each agent only acts on its local observations, when the spatial discretization changes we can simply add more trained agents. 
As shown in Table~\ref{table:euler_test}, the agents have learned the standard WENO agents' policy and the system is able to perform almost exactly like the state-of-the-art WENO numerical scheme for solving Euler equations.

\begin{table}[ht]
	\centering
	\caption{Comparison of trained \textit{RL-WENO} agents' and standard WENO agents' L2 error with the analytical solution for Euler equations on different ICs.}
	\resizebox{\columnwidth}{!}{
		\begin{tabular}{c|c|c|c|c|c|c|c|c}
			\toprule
			IC & \multicolumn{2}{c|}{\textit{Sod}} & \multicolumn{2}{c|}{\textit{Sod2}} & \multicolumn{2}{c|}{\textit{Lax}} & \multicolumn{2}{c}{\textit{Sonic Rarefaction}} \\
			\midrule
			$N$ & \textbf{RL-WENO} & \textbf{WENO} & \textbf{RL-WENO} & \textbf{WENO}& \textbf{RL-WENO} & \textbf{WENO}& \textbf{RL-WENO} & \textbf{WENO}\\
			\midrule
			64  & 0.0707  & 0.0707 & 0.0628 & 0.0628 & 0.5275 & 0.5280 & 2.1192 & 2.1183\\ 
			\midrule
			128 & 0.0420 & 0.0420 & 0.0407 & 0.0407 & 0.4109 & 0.4110 & 1.2401 & 1.2402 \\
			\midrule
			256 & 0.0278 & 0.0278 & 0.0267 & 0.0267 & 0.2411 & 0.2411 & 0.7867 & 0.7866 \\
			\midrule
			512 & 0.0218 & 0.0218 & 0.0183 & 0.0183 & 0.2232 & 0.2233 & 0.5531 & 0.5532 \\
			\bottomrule
		\end{tabular}
		\label{table:euler_test}
	}
\end{table}

\begin{figure}[h]
    \centering
    \begin{subfigure}[b]{0.44\textwidth}
        \centering
        \includegraphics[width=\textwidth]{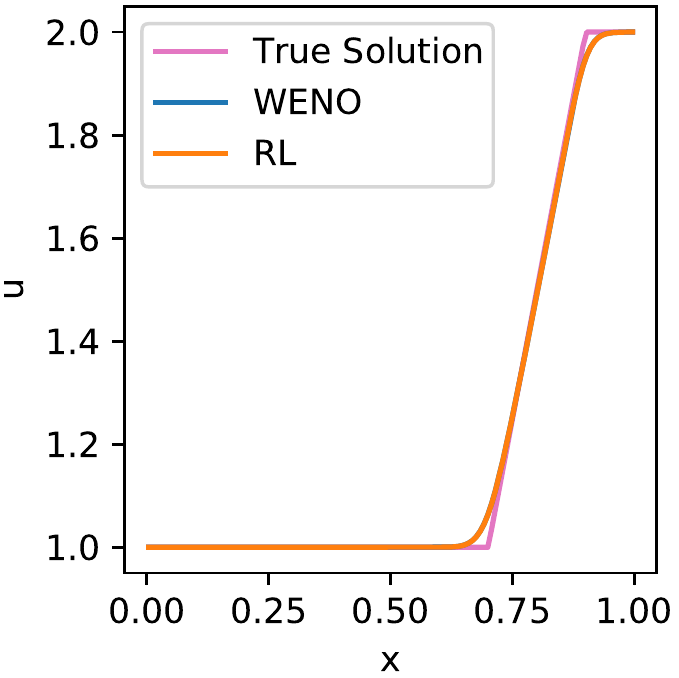}
        \caption{Burger's Equation with rarefaction initial condition, evolved to 0.2s.}
        \label{fig:burgers}
    \end{subfigure}
    \begin{subfigure}[b]{0.55\textwidth}
        \centering
        \includegraphics[width=\textwidth]{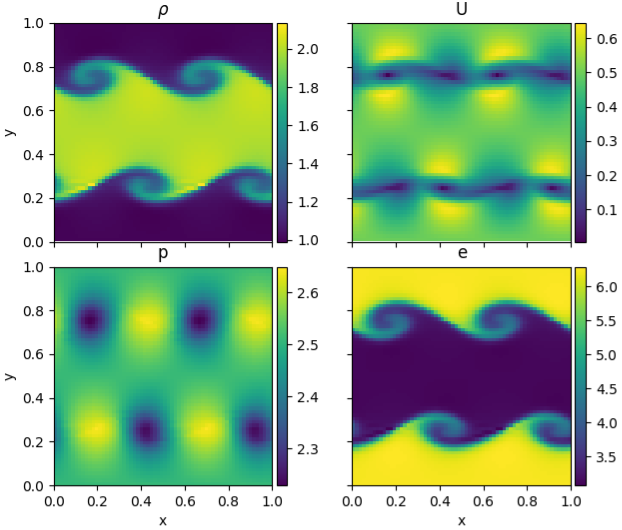}
        \caption{2D Euler Equations with Kelvin-Helmholtz initial condition, evolved to 2.0s.}
        \label{fig:euler_2d:rl}
    \end{subfigure}
        \caption{Testing \textit{RL-WENO} agents on different equation and different dimension.}
        \label{fig:generalization}
\end{figure}

Furthermore, we show that the \textit{RL-WENO} agents trained on 1D Euler Equations can be applied to solving a different hyperbolic PDE, Burger's Equation, as shown in Figure~\ref{fig:burgers}. 
The agents behave exactly like the standard WENO scheme and generate solutions close to the true analytical solution. 
The same agents can also be applied to solving 2D Euler equations as shown in Figure~\ref{fig:euler_2d:rl} by acting on both dimensions. 
These results show that the agents have truly learned the physics and the multi-agent system can perform new tasks.

\section{Conclusion}~\label{sec:conclusion}
In this paper, we introduced both factored Dec-MDP and numerical methods and formulated the learning of WENO scheme as a multi-agent learning problem.
We analyzed different reward formulations, which lead to reinforcement learning (RL) or behavior cloning.
We experimentally tested these formulations and showed that the agents could learn a policy under the RL formulation using a policy gradient algorithm.
Because of the flexibility provided by this factored Dec-MDP framework, the trained agents can be applied to different spatial discretizations, episode lengths, and even equations.
This paper aims to bridge the gap between the domains of scientific computing and multi-agent systems. 
There are many future directions for this work, including but not limited to learning new numerical schemes that go beyond imitating existing methods by automatically discovering better ones.

%
%
\bibliographystyle{splncs04}
\bibliography{PAAMS22_Dec_MDP}

\end{document}